\documentclass[twocolumn,showpacs,preprintnumbers,amsmath,amssymb]{revtex4}

\usepackage{url}
\usepackage{hyperref}
\hypersetup{
    colorlinks,
    citecolor=blue,
    filecolor=black,
    linkcolor=blue,
    urlcolor=blue
}

\usepackage{mathrsfs}
\usepackage{verbatim}
\usepackage{graphicx}	
\usepackage{dcolumn}	
\usepackage{bm}		
\usepackage{bbm,upgreek}
\usepackage{mathtools}
\usepackage{upgreek,xspace}

\usepackage{color}
\usepackage[cal=boondoxo]{mathalfa}
\usepackage{palatino}

\usepackage{dsfont}
\usepackage{cancel}
\usepackage{xcolor}
\usepackage{harpoon}
\usepackage{accents}

\newcommand*\diff{\mathop{}\!\mathrm{d}}

\DeclareFontFamily{U}{euc}{}
\DeclareFontShape{U}{euc}{m}{n}{<-6>eurm5<6-8>eurm7<8->eurm10}{}
\DeclareSymbolFont{AMSc}{U}{euc}{m}{n}
\DeclareMathSymbol{\umu}{\mathord}{AMSc}{"16}

\DeclareSymbolFont{AMSb}{U}{msb}{m}{n}
\DeclareMathSymbol{\N}{\mathbin}{AMSb}{"4E}
\DeclareMathSymbol{\Z}{\mathbin}{AMSb}{"5A}
\DeclareMathSymbol{\R}{\mathbin}{AMSb}{"52}
\DeclareMathSymbol{\Q}{\mathbin}{AMSb}{"51}
\DeclareMathSymbol{\I}{\mathbin}{AMSb}{"49}
\DeclareMathSymbol{\C}{\mathbin}{AMSb}{"43}

\begin{document}

\title{An objective prior that unifies  objective Bayes and information-based inference.}

\author{Colin H. LaMont and Paul A. Wiggins}
\affiliation{Departments of Physics, Bioengineering and Microbiology, University of Washington, Box 351560.\\ 3910 15th Avenue Northeast, Seattle, WA 98195, USA}

\email{pwiggins@uw.edu}\homepage{http://mtshasta.phys.washington.edu/}

\begin{abstract}
There are three principle paradigms of statistical inference: (i) Bayesian, (ii) information-based and (iii) frequentist inference  \cite{Bandyopadhyay2011,Wikipedia4great}. We describe an objective prior (the weighting or $w$-prior) which unifies objective Bayes and information-based inference. The $w$-prior is chosen to make the marginal probability an unbiased estimator of the predictive performance of the model. This definition has several other natural interpretations. From the perspective of the information content of the prior, the $w$-prior is both uniformly and maximally uninformative. The $w$-prior can also be understood to result in a uniform density of distinguishable models in parameter space. Finally we demonstrate the the $w$-prior is equivalent to the Akaike Information Criterion (AIC) for regular models in the asymptotic limit. The $w$-prior appears to be generically applicable to statistical inference and is free of {\it ad hoc} regularization. The mechanism for suppressing complexity is analogous to AIC: model complexity reduces model predictivity. We expect this new objective-Bayes approach to inference to be widely-applicable to machine-learning problems including singular models.
\end{abstract}


\keywords{}

\maketitle

\medskip
\noindent
{\bf Introduction.}
A long-standing goal of statistical inference is the formulation of a consistent and generally-applicable objective-Bayes methodology \cite{Berger2004}. Although the Bayes law formally depends on knowledge of a prior probability distribution (prior), from its infancy, Bayesian analysis has been applied in scenarios where this prior information is unknown \cite{Jeffreys1939,Laplace1812}. 
A second principle paradigm of inference has been developed around a frequentist formulation of probability, beginning with the work of C.~F.~Gauss, R.~A.~Fisher, etc \cite{cox2005}. In the 1970s, a third paradigm of information-based inference was developed based on the pioneering work of Akaike \cite{akaike1773,BurnhamBook,wiki:AIC}. Although it is often possible to arrive at similar conclusions using these distinct statistical paradigms \cite{Bayarri2004}, this is not always the case \cite{Jeffreys1939,wiki:LindleysP}. 



\medskip
\noindent
{\bf Summary.} We motivate the form of the $w$-prior using the {\it Principle of Indifference}. We propose a precise implicit definition of the $w$-prior by defining a {\it Bayes predictive  multiplicity}: the number of indistinguishable models in the vicinity of parameterization $\bm \theta$. The $w$-prior has unit multiplicity for all model parameterizations. We demonstrate that the resulting Bayes partition function is an unbiased estimator for the predictive performance of the learning machine. Next we show that the multiplicity can be understood as the parameter-coding information and 
using this formulation, we demonstrate that the $w$-prior is both uniformly and maximally uninformative. 

Having established that the $w$-prior has many of the desired properties of an objective prior, we explore the connection to the information-based paradigm of statistical inference. We demonstrate that for {\it regular} models \footnote{A model is regular if the eigenvalues of the Fisher Information Matrix are large enough to make the log-likelihood effectively harmonic in the parameter values around their MLE or MAP value.},  objective-Bayes inference is asymptotically equivalent to the Akaike Information Criterion (AIC)  \cite{akaike1773,BurnhamBook,wiki:AIC}, unifying objective-Bayes and information-based inference. 
Finally, we discuss {\it predictivity} as the unifying principle that is common to all three principle paradigms of inference.


\medskip
\noindent
{\bf Preliminaries.} We introduce the following notation for a set of $N$ independent and identically distributed observations \footnote{When $X$ appears in upper case, it should be understood as a random variable whereas it is a normal variable when it appears in lower case. If we need  a statistically independent set of variables of equal size, we will use the random variables $Y^N$, which have identical properties to the $X^N$.}:
\begin{equation}
X^N \equiv (X_1,X_2,...,X_N)\ \ \ {\rm where}\ \ \ X_i \sim q(\cdot|{\bm \theta}_0), 
\end{equation}
and the distribution function $q$ is parameterized by ${\bm \theta}_0$, the true parameterization. We  model the probability distribution with a set of model parameterizations ${\bm \theta}\in{\bm \Theta}$. In general this set will include models of different dimension  (i.e.~{\it complexity} $K\equiv\dim \bm \theta$).
We assume a frequentist realization of Bayesian statistics: There are many realizations of the system of interest and the true parameterization is a random variable with a distribution defined by the true prior distribution ${\bm \theta}_0 \sim \varpi_0(\cdot)$. This true prior may or may not be known.   


We introduce a generalizated marginal probability called the {\it Bayes partition function} \cite{watanabe2009}:
\begin{equation}
Z(X^N|\rho)  \equiv  \int_{\bm \Theta}\!\!\! \diff {\bm \theta}\ \rho({\bm \theta})\,q(X^N|{\bm \theta}), \label{Eqn:BayesPart}
\end{equation}
where $\rho({\bm \theta})$ is a density of models on the manifold $\bm \Theta$ that need not be normalized. When $\rho = \varpi_0$, the marginal probability (partition function) has the meaning probability of observing $X^N$ for an unknown true parameterization.

The generalized {\it posterior probability} is defined:
\begin{equation}
\rho( {\bm \theta} | X^N ) \equiv { \frac{q(X^N|{\bm \theta})\,\rho({\bm \theta})}{Z(X^N|\rho)}}, 
\end{equation}
which is known as the Bayes Law if $\rho=\varpi_0$. Like the partition function, we generalize the posterior to permit updating of any distribution on $\bm \Theta$.  In this generalized case, the {\it posterior} is understood to have the meaning of a model weighting. Only when the posterior is constructed with the true prior is the posterior understood as the probability distribution for the unknown true parameterization, updated by the observations $X^N$. Note that even if the prior is initially improper (unnormalized), the posterior will be normalized.

The Bayes predictive distribution is defined:
\begin{equation}
q( X | X^N,\rho ) \equiv {\frac{Z(X,X^N|\rho)}{Z(X^N|\rho)}},\label{Eqn:inf}
\end{equation}
where $X \notin X^N$ and is understood to be the Bayesian predicted probability distribution for a new observation $X$ given $N$ observations $X^N$ and a prior $\rho$.

We define the {\it Bayesian free energy} \cite{watanabe2009}: 
\begin{equation}
G(X^N,\rho) \equiv -\log Z(X^N|\rho)
\end{equation}
and the closely related {\it performance estimator}: 
\begin{eqnarray}
{\cal P}^N_{\rm post}({\bm \theta},\rho)  &\equiv& \underset{q(\cdot|{\bm \theta})}{\mathbb{E}_{X}} \ \log Z(X^N|{\rho}).\label{Eqn:NPerf}
\end{eqnarray}
The significance of this definition and its interpretation as an estimator will be discussed shortly. 

Finally we introduce the natural measure of predictive performance: the ability of the trained Bayes model to predict new observations. Again it is convenient to formulate the performance of the model for $N$ new observations. We define the {\it predictivity}:
\begin{eqnarray}
{\cal P}_{\rm pre}^N({\bm\theta},\rho) &\equiv& N\ \underset{q(\cdot|{\bm \theta})}{\mathbb{E}_{X}} \log q(X|X^N,\rho), \label{Eqn:NPred}
\end{eqnarray}
where $X \notin X^N$. The predictivity is the Bayesian predictive performance for $N$ simultaneous measurements. 

\medskip
\noindent
{\bf Principle of Indifference.}
In instances where no prior information was known, both P.~S.~Laplace \cite{Laplace1812} and T.~Bayes \cite{Bayes1763} invoked a {\it principle of indifference} which assigned {\it mutually exclusive} and {\it exhaustive} possibilities equal prior probability \cite{Keynes1921,PrincipleOfIndifference}. Although this approach does lead to consistent results in the context of models with discrete parameterization, it has long been unclear how to generalize the principle of indifference to a continuum context where the meaning of both  {mutually exclusive} and {exhaustive} are uncertain. 


\begin{figure}
  \centering
    \includegraphics[width=0.5\textwidth]{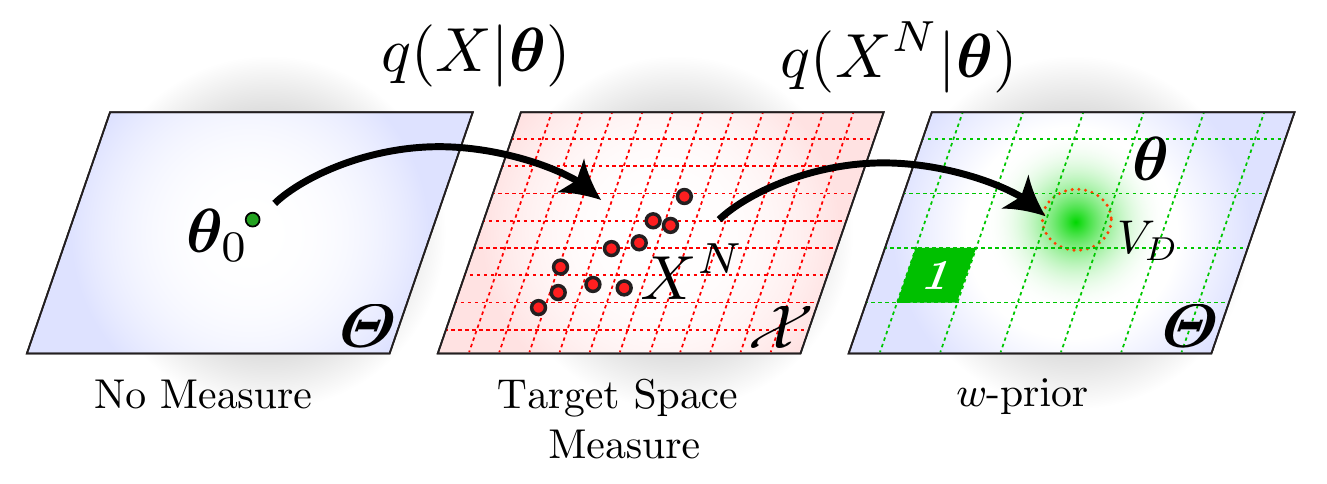}
      \caption{{\bf Geometry of inference. Panel A: Inference defines a natural measure on parameter space $\bm \Theta$.} To understand the origin of this measure, consider a true parameterization ${\bm \theta}_0$ used to generate observations $X^N$. We then define a cutoff divergence $D_{\rm D}$ to determine a region of indistinguishability on $\bm \Theta$ with volume $V_{\rm D}({\bm \theta})$. We define the density of models as $\rho_{\rm D} \equiv V_{\rm D}^{-1}$.  
          \label{Fig:GeoInf}}
\end{figure}

\medskip
\noindent
{\bf The prior depends on $N$.} In the interest specificity, consider the set of gaussian distributions with mean $\mu \in \mathbb{R}$ and variance $v=1$. We can write ${\bm \theta} \equiv (\mu,v)$. Are  two models with $\mu_1 \ne \mu_2$ mutually exclusive? Intuitively we know the answer: If the difference in the means is sufficiently small, we cannot distinguish the models for a small number of observations. But, whatever the values of the means $\mu_i$, if $\mu_1 \ne \mu_2$ then a sufficient number of observations $N$ can always resolve the models, rendering them mutually exclusive. Therefore it is natural to assume that the condition for model distinguishability  depends on the number of observations.

\medskip
\noindent
{\bf Estimating the density of models.} V.~Balasubramanian and others \cite{Balasubramanian1997} have argued  that a natural measure for testing distinguishability for ${\bm \theta}_1 \ne {\bm \theta}_2$ exists and is given by the KL Divergence \cite{Kullback1951a,wiki:KLD}:
\begin{equation}
D({\bm \theta}_1||{\bm \theta}_2) \equiv \int_{\cal X}\!\! {\rm d} x\ q(x|{\bm \theta}_1)\ \log {\frac{q(x|{\bm \theta}_1)}{q(x|{\bm \theta}_2)}},
\end{equation}
which is a  measure of distance between any two probability distributions. For ${\bm \theta}_1={\bm \theta}_2$, the divergence is identically zero and $D>0$ for $q(x|{\bm \theta}_1) \ne q(x|{\bm \theta}_2)$ \cite{Kullback1951a,wiki:KLD}. 

For simplicity, assume that we pick some critical value of the divergence $D_{\rm D}$, below which two distributions are considered  indistinguishable ($\sim_{\rm D}$) and above which two distributions are distinguishable ($\nsim_{\rm D}$ or {\it mutually exclusive}). The sum over equally weighted mutually exclusive distributions can be written as an integral \cite{Balasubramanian1997}:
\begin{equation}
\sum_{\nsim_{\rm D}} \longrightarrow  \int_{\bm \Theta}\!\! \diff {\bm \theta}\ \rho_{\rm D}({\bm \theta}),\label{Eqn:theSum}
\end{equation}
where $\rho^{-1}_{\rm D}=V_{\rm D}$ is the parameter-space volume of parameterizations that are equivalent in the vicinity of ${\bm \theta}$. (For the moment, assume all ${\bm \theta} \in \bm \Theta$ are continuous parameters of dimension $K$. A schematic drawing of the meaning of the parameter-space volume is illustrated in Fig.~\ref{Fig:GeoInf}.) The divergence for $N$ observations is simply $N$ times the divergence for one.  For a large number of observations $N$ and a regular model, we can write the density of distinguishable model as \cite{Balasubramanian1997}:
\begin{equation}
\rho_{\rm D} \propto  \sqrt{\det {\bm I}}\ \left({\textstyle \frac{N}{2\pi}} \right)^{K/2}\ {D_{\rm D}^{-K/2}}, \label{eqn:dospi}
\end{equation}
where ${\bm I}$ is the Fisher Information Matrix \cite{FIM}. Clearly this density is increasing in $N$ since the resolution of the learning machine increases with the number observations. The canonical approach is to drop the $N$ dependence, keeping only the determinant in the Fisher Information Matrix \cite{Balasubramanian1997}.
But for the purpose of counting distinguishable models with distinct dimension $K$, it is essential to retain the $N$ dependence since it cannot be factored out of the sum in Eqn.~\ref{Eqn:theSum}.

We will make a more precise definition of the density of models shortly, but the insistence that the prior (the density of models) depends on the number of observations $N$ has profound consequences as we describe below. 

Note that for fixed model dimension $K$, we will show that Eqn.~\ref{eqn:dospi} is correct for a specific value of the critical divergence $D_{\rm D}$.
 One might hope that inference was either independent or weakly dependent on $D_{\rm D}$ but in fact its value is critically important. We therefore need to propose a more precise definition of the density of models.



\medskip
\noindent
{\bf The  multiplicity and the $w$-prior.} To define the density of models precisely, we first introduce the Bayes predictive  multiplicity $m$:
\begin{equation}
\log m({\bm \theta}_0,\rho) \equiv {\cal P}^N_{\rm post}({\bm \theta}_0,\rho)-{\cal P}_{\rm pre}^N({\bm \theta}_0,\rho). \label{Eqn:MeaningOfPart} 
\end{equation}
The meaning of the multiplicity is the number of indistinguishable models defined by the prior in the vicinity of ${\bm \theta}_0$. (We will return to this discussion shortly.)

We define the weighting prior ($w$-prior) as the density of indistinguishable models such that the multiplicity is unity for all true models in the set ${\bm \Theta}$:
\begin{equation}
m({\bm \theta}_0,w) = 1\ \ \ \ \forall \ \ \ \ {\bm \theta}_0 \in {\bm \Theta}. \label{Eqn:wprior}
\end{equation}
Clearly $w$ will in general be an improper prior. The name ``weighting'' is chosen to emphasize the {\it model-weighting} rather than probabilistic interpretation of analysis as described below \cite{BurnhamBook}.

\medskip
\noindent
{\bf An unbiased estimator of performance.} For an improper prior, the  partition funciton can no longer be understood as the marginal probability of observations $X^N$. But the definition of the $w$-prior gives the partition function a precise mathematical meaning: Eqn.~\ref{Eqn:MeaningOfPart}, evaluated at the $w$-prior (Eqn.~\ref{Eqn:wprior}), implies that the performance estimator (Eqn.~\ref{Eqn:NPerf}) and by extension the log partition function (Eqn.~\ref{Eqn:BayesPart}) are  {\it unbiased estimators} \cite{BurnhamBook,wiki:Bias} of the predictive performance of the model (Eqn.~\ref{Eqn:NPred}). Therefore the partition function is understood as the probability of the last $N$ measurements in the subjective-Bayes framework, but as an estimate of the probability of the next $N$ in the objective-Bayes framework.

\medskip
\noindent
{\bf Connection to the Gibbs Entropy.} To understand the mathematical meaning of the multiplicity $m$ (Eqn.~\ref{Eqn:MeaningOfPart}), it is helpful to transform the equation to write it in terms of the posterior probability distribution. We analytically continue the number of observations to define the continuous variable {\it effective temperature}: $T\equiv N^{-1}$ (e.g.~\cite{watanabe2009}). We can now identify the  definition of the multiplicity as exactly analogous to the computation of the disorder-averaged Gibbs entropy ($\overline{S}$) (e.g.~\cite{Reif1967,wiki:Gibbs}):
\begin{eqnarray}
\overline{S}(T,\rho,{\bm \theta}_0) \equiv -\partial_T \overline{F} = \log m + {\cal O}(T), \label{Eqn:statmechInf}
\end{eqnarray}
from the disorder-averaged Helmholtz free energy ($\overline{F}$):
\begin{eqnarray}
\overline{F}(T,\rho,{\bm \theta}_0) \equiv -T\left< \log Z\right>_X,
\end{eqnarray}
where the angle brackets represent the expectation over $X^N$ with respect to the true distribution parameterized by ${\bm \theta}_0$ and the order $T$ correction is an error due to the analytic continuation of the number of observations $N$ \cite{wPriorLong}.   In a physical context, $X^N$ is  quenched disorder \cite{Balasubramanian1997}. The $w$-prior satisfies the condition that the disorder-averaged Gibbs entropy is  zero to order ${\textstyle \frac{1}{N}}$:
\begin{equation}
\overline{S}({\textstyle \frac{1}{N}},w,{\bm \theta}_0) = 0 + {\cal O}({\textstyle \frac{1}{N}}), \label{Eqn:GibbsZero}
\end{equation}
 for all ${\bm \theta}_0 \in {\bm \Theta}$. 

\medskip
\noindent
{\bf The meaning of multiplicity.} Re-expressing Eqn.~\ref{Eqn:statmechInf} in statistical quantities gives the following expression for the multiplicity: 
\begin{equation}
\log m({\bm \theta}_0,\rho) =  \underset{q(\cdot|{\bm \theta}_0 )}{\mathbb{E}_{X,Y} }\ \ \underset{\rho(\cdot|Y^N)}{\mathbb{E}_{\bm \theta} } {\textstyle \log \frac{ \rho( {\bm \theta} )}{  \rho( {\bm \theta}|X^N )}} + {\cal O}({\textstyle \frac{1}{N}}). \label{Eqn:mult2}
\end{equation}
The meaning of the multiplicity is understood as follows: ${\bm \theta}$ can be understood as the estimator of ${\bm \theta}_0$. $\rho( {\bm \theta} )$ is the density of models at ${\bm \theta}$ and $V_{\rm D}\approx \rho^{-1}( {\bm \theta}|X^N )$ is the parameter-space volume of indistinguishable models.  Therefore the multiplicity $m$ is the number of indistinguishable models defined by the prior $\rho( {\bm \theta} )$. See Fig.~\ref{Fig:GeoInf} for a schematic illustration.

Although this expression is almost in the form of the divergence, the expectation is taken over the posterior probability distribution for an independent dataset $Y^N$. This {\it cross-validation} form is a key distinction and avoids the over-fitting phenomena \footnote{Note that this cross-validation is {\it virtual}. No explicit cross-validation will be performed on true data.}. 

\medskip
\noindent
{\bf The parameter-coding interpretation.} We now wish to re-interpret the multiplicity as a parameter-coding information. We define the parameter-coding information content of the observations:
\begin{eqnarray} 
\overline{H}_{\bm \theta}({\bm \theta}_0,\varpi) \equiv \underset{q(\cdot|{\bm \theta}_0 )}{\mathbb{E}_{X,Y} }\ \  \underset{\varpi(\cdot|X^N)}{\mathbb{E}_{\bm \theta} } \log {\textstyle \frac{\varpi({\bm \theta}|Y^N) }{\varpi({\bm\theta})}},\label{Eqn:ParamEncode}
\end{eqnarray}
given a normalized prior $\varpi$. Note that as before we are careful to define the parameter-coding information by {\it cross-validation} as before with posterior probability distributions generated from independent datasets $X^N$ and $Y^N$.

We now need to define a normalized $w$-prior. We integrate overall parameter space (applying a cutoff if necessary) to determine the normalization constant (total number of models):
\begin{eqnarray} 
N_{\bm \Theta} \equiv \int_{\bm \Theta} \!\! {\rm d} {\bm \theta}\ w({\bm \theta}).
\end{eqnarray}
We define the normalized $w$-prior: $\varpi_w \equiv N^{-1}_{\bm \Theta} w$ \footnote{Note that regardless of whether $N_{\bm \theta}$ is strictly convergent, the multiplicity is well defined. Only our interpretation as a parameter-coding information is dependent of regularizing $N_{\bm \Theta}$.}. Inserting the normalized $w$-prior into the  parameter-coding information gives
\begin{eqnarray} 
\overline{H}_{\bm \theta}({\bm \theta}_0,\varpi_w) = \log N_{\bm \Theta} +{\cal O}({\textstyle \frac{1}{N}}),
\end{eqnarray}
which is clearly constant with respect to the true model ${\bm \theta}_0$ up to order ${\textstyle \frac{1}{N}}$.


\medskip
\noindent
{\bf The $w$-prior is uniformly uninformative.}  We now wish to analyze the dependence of the average parameter-coding information $\overline{H}_{\bm \theta}$ on  the  true model ${\bm \theta}_0$. Informative priors have the property that they code information about the underlying true model. A prior localized around the true value ${\bm \theta}_0$ will reduce the information content of the observations. Therefore we might intuitively expect an uninformative prior to result in a constant information content with respect to the true model ${\bm \theta}_0$. The $w$-prior has  this property to order ${\textstyle \frac{1}{N}}$. No parameter values are favored or disfavored.  We therefore say that the $w$-prior is {\it uniformly uninformative} since the information gain is independent of the true model. 

\medskip
\noindent
{\bf The $w$-prior is maximally uninformative.} In the definition of the reference prior, J.~M.~Bernardo argued that  an {\it objective} and {\it uninformative} prior should maximize the information specified by the observations \cite{BernardoBerger1991}.  We therefore construct an expression analogous to that which he proposed, differing only in the use of the cross-validation form of the parameter-code information: 
\begin{eqnarray} 
\overline{\overline{H}}_{\bm \theta}(\varpi) \equiv \underset{\varpi}{\mathbb{E}_{\bm \theta} } \ \overline{H}_{\bm \theta}({\bm \theta},\varpi),
\end{eqnarray}
which can be understood as the expectation of the $\overline{H}_{\bm \theta}({\bm \theta},\varpi)$ over a true prior $\varpi$. A prior which is maximally uninformative, should be stationary with respect to variations in the prior. $\varpi_w$ is stationary and a least locally a maximum to order $N^{-1}$. We therefore conclude that the $w$-prior also possesses the property that it is {\it maximally uninformative}.

\medskip
\noindent
{\bf The $w$-prior for a regular model.}
 In general, there is no closed-form expression of the $w$-prior although it can be computed exactly for a number of special cases and for sufficiently simple models \cite{wPriorLong}.  
 
In the interest of simplicity consider  a regular model where the number of continuous degrees of freedom in the model parameterization $\bm \theta$ is $K$ and work in the large number of observations $N$ limit. Using the Gibbs entropy equation for the $w$-prior, we compute the $w$-prior \cite{wPriorLong}:  
\begin{eqnarray}
w({\bm \theta}) &=& J\ \left({\textstyle \frac{N}{2 \uppi}}\right)^{K/2}\ e^{-K}, \label{Eqn:wpriorReg}
\end{eqnarray}
where  
\begin{eqnarray}
J({\bm \theta})  &\equiv &  \sqrt{\det {\bm I}\,},
\end{eqnarray}
is equal to the square root of the determinant of the Fisher Information Matrix and is the well known Jeffreys prior that H.~Jeffreys proposed to insure invariance of the probability to re-parameterization of $\bm \theta$ \cite{Jeffreys1946,wiki:JP}. It is instructive to immediately compare this results to form we estimated based on principle of indifference (Eqn.~\ref{eqn:dospi}). These arguments correctly identified both the Jeffreys prior factor $J$ and the scaling with the number of observations $N$. But, a precise formulation was required to correctly compute the last factor in Eqn.~\ref{Eqn:wpriorReg}, the penalty $e^{-K}$, which plays a critical role in the regularization of the $w$-prior when the complexity of the model is unknown. 

Clearly Eqn.~\ref{Eqn:wpriorReg} implies that for any significant number of observations $N$, the $w$-prior appears to increase with model complexity $K$, but all factors except the for $e^{-K}$ cancel during marginalization over $\bm \theta$ in the computation of the partition function (Eqn.~\ref{Eqn:BayesPart}). The qualitative understanding of the $w$-prior is therefore a penalization (regularization) of model complexity.

\medskip
\noindent
{\bf Equivalence to information-based inference.}  We now compute the $w$-prior for a regular model of unknown complexity $K$. Typically for models of unknown complexity, the $w$-prior cannot be computed exactly. We have therefore developed a recursive technique  \cite{wPriorLong} analogous to that proposed by J.~M.~Bernardo \cite{BernardoBerger1991}. 
We write the total model parameterization as ${\bm \theta}\equiv (K,{\bm \theta}^K)$ where the ${\bm \theta}^K$ are $K$ continuous parameters. The first-order expression for the $w$-prior for a regular model of unknown complexity is still given by Eqn.~\ref{Eqn:wpriorReg} with ${\bm \theta} \rightarrow {\bm \theta}^K$ \cite{wPriorLong}. The partition function using the first-order expression for the $w$-prior is:
\begin{eqnarray}
Z &=& \sum_{K=1}^\infty Z_K, \\
G_K &\equiv& -\log Z_K = -\log q( X^N| K, \hat{\bm \theta}_X^K) + K,
\end{eqnarray}
where the  $\hat{\bm \theta}_X^K$ are the Maximum Likelihood Estimators of the parameters $\bm \theta^K$, $Z_K$ and $G_K$ are the partition function and free energy at complexity $K$. The first term in the free energy is the minus-log likelihood and the second term is interpreted as a penalization for model complexity. To those familiar with the information-based approach of Akaike, it is clear that the free energy $G_K$
is identical to AIC:
\begin{equation}
G_K = {\rm AIC}_K,
\end{equation}
for a model with $K$ degrees of freedom \cite{akaike1773,BurnhamBook}. Therefore, in the asymptotic limit for regular models, the $w$-prior will simply recover information-based inference \footnote{In fact in the context of discussing the connection between the Bayesian Information Criterion (BIC) \cite{BurnhamBook,Schwarz1978,wiki:BIC} and AIC, K.~Burnham and D.~Anderson suggested that just such a prior would be a {\it savvy prior} due to its good frequentist attributes \cite{BURNHAM2004}.}. The $K$ penalty is the information-based realization of {\it Occam's Razor}: parsimony implies predictivity. 

\medskip
\noindent
{\bf The $w$-prior for singular models.} Singular models contain parameters for which the Fisher Information is zero (or nearly zero for finite $N$). AIC fails in the context of singular models but we have recently proposed a generalization of AIC called the Frequentist Information Criterion (FIC) for application to singular models \cite{FICshort,FICapp}, which is equivalent to Neyman-Pearson hypothesis testing \cite{FICshort,FICapp}. Given the close connection between AIC and the $w$-prior described above, one might hope that $w$-prior for a singular model would be analogous to FIC. The equivalence between  information-based inference and the $w$-prior appears to a hold only for regular models in the asymptotic limit ($N\rightarrow \infty$). It is well known that  Bayesian inference, corresponding to the Schwartz distribution topology, has a finite generalization error in the asymptotic limit, whereas maximum-likelihood-based techniques result in divergent generalization errors in the asymptotic limit (e.g.~\cite{watanabe2009}). We provide a detailed description of the connections between objective bayes, information-based and frequentist inference elsewhere \cite{FICshort,wPriorLong}. 

%
%


\medskip
\noindent
{\bf The weighting interpretation.} The $w$-prior and posterior probability distribution for the model parameterization ${\bm \theta}$ should be understood as a  model {\it weighting}, not as the probability density that ${\bm \theta}$ is the true parameterization ${\bm \theta}_0$. We have given the $w$-prior the name of {\it weighting} prior in close analogy to the Akaike weights (e.g.~\cite{BurnhamBook}). From a frequentist perspective, we are forbidden from discussing the probability of a model which we cannot compute since we do not know the prior from which the truth was constructed. The {\it weighting} interpretation is not simply philosophical point, but has important computational significance.  For instance, as the number of observations $N$ increases, the resolution of the objective-Bayes learning machine increases also and therefore as a consequence the {\it weighting} of more complex models in the $w$-prior increases also. As a result, the complexity of the fit model naturally increases with the size of the dataset when describing an infinite dimensional true model, as predicted by the information-based approach. The infinite complexity of the true model is of no consequence to the selection of the objective $w$-prior. 

Like  subjective-Bayes inference, the $w$-prior generates a weighted ensemble of models rather than a point estimate or confidence intervals and therefore has  both the associated advantages and short-comings of the Bayesian machinery. A number of authors (e.g.~\cite{Loredo1990}) have argued that the frequentist approach is itself {\it ad hoc} due to the bewildering proliferation of tests and statistics. These authors may view the $w$-prior, not as a tool for Bayesian statisticians, but rather as a missing unifying principle for frequentist methods to place them on par with existing Bayesian methods.

\medskip
\noindent
{\bf Cross-validation.} Predictivity, cross-validation, bootstrapping and generalization error are all essentially mathematically equivalent measures of model performance \footnote{These measures are equivalent for a large number of observations $N$.}. Therefore, clearly the $w$-prior can be interpreted to be weighted to optimize cross-validation or generalization-error-based measures of performance. In fact, it has recently been formally demonstrated that stability to cross-validation is a necessary and sufficient condition for the predictive performance of a learning machine \cite{Poggio2004}.  

\medskip
\noindent
{\bf The central role of predictivity.} Motivated by the work of Akaike, we have repeatedly made use of the principles of {\it predictive performance}. For instance, see Eqns.~\ref{Eqn:MeaningOfPart}, \ref{Eqn:ParamEncode} and \ref{Eqn:SubBayesPred}. The critical consideration in each of these equations is always {\it generalization}: the model performance measured against data not included in the training set. It is this formulation that leads to {\it model selection} (or {\it model regularization}). For instance, if one were to define the multiplicity (Eqn.~\ref{Eqn:MeaningOfPart}) with respect to the postdictive performance \footnote{Postdiction: The performance of the model in reproducing the training set.} or even the performance of the true model \footnote{J.-M.~Bernardo has proposed a somewhat analogous prior, the reference prior, which uses the true model as the reference in the multiplicity (e.g.~\cite{BernardoBerger1991}). This approach has exactly the shortcomings one would predict. The prior is not consistent for nested models.}, the $w$-prior could not be consistently defined for an infinitely-nestable model since there would be an ultraviolet divergence \cite{UVdiv} for high-complexity (large $K$) models. It is  the use of the predictive performance as the model weighting that gives rise to regularization which is both natural and statistically-principled. It is unnecessary to augment the predictive regularization with exogenous  and {\it ad hoc} regulatory devices such as smoothing \cite{MacKay1992a}, hyper-parameters \cite{MacKay2003} or vague priors \cite{Kass1995}.

\medskip
\noindent
{\bf The maximum-predictivity interpretation.} Finally we wish to discuss our results in the context of subjective Bayes analysis where the true prior is known. We note the true prior is optimal in two senses. (i) The true prior maximizes the expectation of the performance estimator (e.g.~\cite{watanabe2009}):
\begin{equation}
\overline{\cal P}^N_{\rm post}(\varpi_0,\varpi)  \equiv \underset{\varpi_0}{\mathbb{E}_{\bm \theta}} \ {\cal P}^N_{\rm post}({\bm \theta},\varpi),
\end{equation}
which has a unique global maximum at $\varpi=\varpi_0$. But with particular relevance to our current work, the true prior is also optimal in a predictive sense. (ii) The true prior maximizes the expectation of the predictivity (e.g.~\cite{watanabe2009}):
\begin{equation}
\overline{\cal P}^N_{\rm pre}(\varpi_0,\varpi)  \equiv \underset{\varpi_0}{\mathbb{E}_{\bm \theta}} \ {\cal P}^N_{\rm pre}({\bm \theta},\varpi),\label{Eqn:SubBayesPred}
\end{equation}
which has a unique global maximum at at $\varpi=\varpi_0$. It is tempting ask whether there is some prior that optimizes predictivity directly if the true prior is not know, in analogy to Eqn.~\ref{Eqn:SubBayesPred}, but no such prior exists \cite{watanabe2009,wPriorLong}.

\medskip
\noindent
{\bf Discussion.} We have presented predictive performance as a unified framework for reconsiling the three principle paradigms of statistical inference \cite{Bandyopadhyay2011,Wikipedia4great}. As discussed in the previous section, subjective-Bayesian analysis can be understood to directly maximize the predictive performance of the model if (and only if) the true prior is used to generate inference. 

But by far the most important practical scenario is an unknown true prior. In this cases, the predictive performance of the model cannot be strictly maximized since the true prior is unknown. In this scenario, we propose that inference be performed by weighting models by their expected predictive performance using the $w$-prior. As we have demonstrated, the $w$-prior results in a partition function which is the unbiased estimator of model performance. Therefore inference using the $w$-prior can also be understood as the optimization of predictive performance, although not in the strict sense of maximization.

The $w$-prior is improper and yet has a rigorous statistical meaning. There has been a long history of the successful application of improper priors in statistical analysis, most famously by H.~Jeffreys \cite{Jeffreys1946}, and despite considerable, ongoing and heated debated about the statistical meaning and rigor of these approaches \cite{Dawid1973,Dawid1980,Jaynes2003,MacKay2003}. We have proposed one possible rigorous definition for an improper prior and have described why such priors have good performance from a frequentist perspective.

The $w$-prior also has a natural interpretation as an uninformative prior. (i) It is a realization of the {\it Principle of Indifference} in the sense that the prior weights all distinguishable models equally. Since this scenario describes a state of maximum entropy, the $w$-prior has a MaxEnt interpretation. (ii) It is {\it uniformly uninformative} in the sense the parameter-information content of the observations is independent of the true model. (iii) It is   {\it maximally uninformative} in the sense that it maximizes the model-averaged parameter-information content of observations and can therefore be interpreted as a {\it reference prior}. Finally we demonstrated that the $w$-prior is equivalent to information-based AIC inference for regular models.
The $w$-prior has virtually all of the desirable properties of an objective-Bayesian prior with one key short-coming: Both the prior and the posterior have a {\it weighting} rather than a Bayesian probabilistic interpretation. We believe that such an interpretation is not only philosophically desirable but a mathematical necessity.  



\bibstyle{IEEEtran}

\bibliography{../bib/ModelSelection}

\end{document}